\documentclass{article}
\usepackage{csquotes}
\usepackage{spconf,amsmath,graphicx,siunitx,microtype,amsfonts}
\usepackage[numbers]{natbib}
\usepackage{array}
\usepackage{booktabs}
\usepackage{microtype}
\usepackage{url}
\usepackage{amsmath}
\usepackage{siunitx}
\usepackage{color}
\usepackage{multirow}
\usepackage{booktabs}
\usepackage{float}
\usepackage{bbding}

\ninept

\title{Time-aware Multiway Adaptive Fusion Network for Temporal Knowledge Graph Question Answering}
\name{Yonghao Liu$^{\Diamond\clubsuit}$, Di Liang$^{\Diamond\clubsuit}$, Fang Fang$^{\Diamond\spadesuit}$\thanks{$^{\Diamond}$ Equal contribution.} 
,Sirui Wang$^{\spadesuit}$
,Wei Wu$^{\clubsuit}$,Rui Jiang$^{\spadesuit}$
}
\address{
$^\spadesuit$ Department of Automation, Tsinghua University, Beijing, China \\
			$^\clubsuit$Centre for Natural Language Processing, Meituan Inc., Beijing, China  \\ \{liuyonghao, liangdi04, wangsirui, wuwei30\}@meituan.com  \\\{ff20, ruijiang\}@tsinghua.edu.cn
			}

\begin{document}
%

\maketitle

\begin{abstract}
Knowledge graphs (KGs) have received increasing attention due to its wide applications on natural language processing. However, its use case on temporal question answering (QA) has not been well-explored. Most of existing methods are developed based on pre-trained language models, which might not be capable to learn \emph{temporal-specific} presentations of entities in terms of temporal KGQA task. To alleviate this problem, we propose a novel \textbf{T}ime-aware \textbf{M}ultiway \textbf{A}daptive (\textbf{TMA}) fusion network.
Inspired by the step-by-step reasoning behavior of humans. For each given question, TMA first extracts the relevant concepts from the KG, and then feeds them into a multiway adaptive module to produce a \emph{temporal-specific} representation of the question. This representation can be incorporated with the pre-trained KG embedding to generate the final prediction. 
Empirical results verify that the proposed model achieves better performance than the state-of-the-art models in the benchmark dataset. Notably, the Hits@1 and Hits@10 results of TMA on the CronQuestions dataset's complex questions are absolutely improved by 24\% and 10\% compared to the best-performing baseline. Furthermore, we also show that TMA employing an adaptive fusion mechanism can provide interpretability by analyzing the proportion of information in question representations.

\end{abstract}
\begin{keywords}
 Temporal knowledge graph question answering, knowledge graph question answering, neural language processing
\end{keywords}
\section{Introduction}
\label{sec:intro}
Knowledge graph question answering (KGQA) is a core technique in many natural language processing applications, such as search and recommendation \cite{huang2019knowledge,xian2019reinforcement,liu2021vpalg}. 
Among several branches of KGQA, temporal KGQA is a recently emerging direction and has shown great potential in real-world practices. We note that there are critical differences between traditional KGQA and temporal KGQA tasks, which are summarized as follows:
(I) Temporal KGQA has more complex semantic information, unlike the traditional KGs constructed based on the tuple of (subject, predicate, object)\footnote{Some researchers refer to predicates as relations. The two are equivalent.}, temporal KGs are attached with additional timestamps.
In other words, the tuple of temporal KGs is (subject, predicate, object, time duration). One example is (Barack Obama, position held, President of USA, 2008, 2016) representing that Barack Obama held the position of President of the USA from 2008 to 2016. (II) Temporal KGQA is expected to generate answers with more diverse types. Different from regular KGQA whose answers are always entities, the answer of temporal KGQA can either be an entity (\textit{e.g.}, Barack Obama) or a timestamp (\textit{e.g.}, 2008, 2016). The above differences make the temporal KGQA much more challenging to solve, as it often requires additional temporal reasoning compared to traditional KGQA tasks. 

\begin{figure}
\centering
\includegraphics[width=0.48\textwidth]{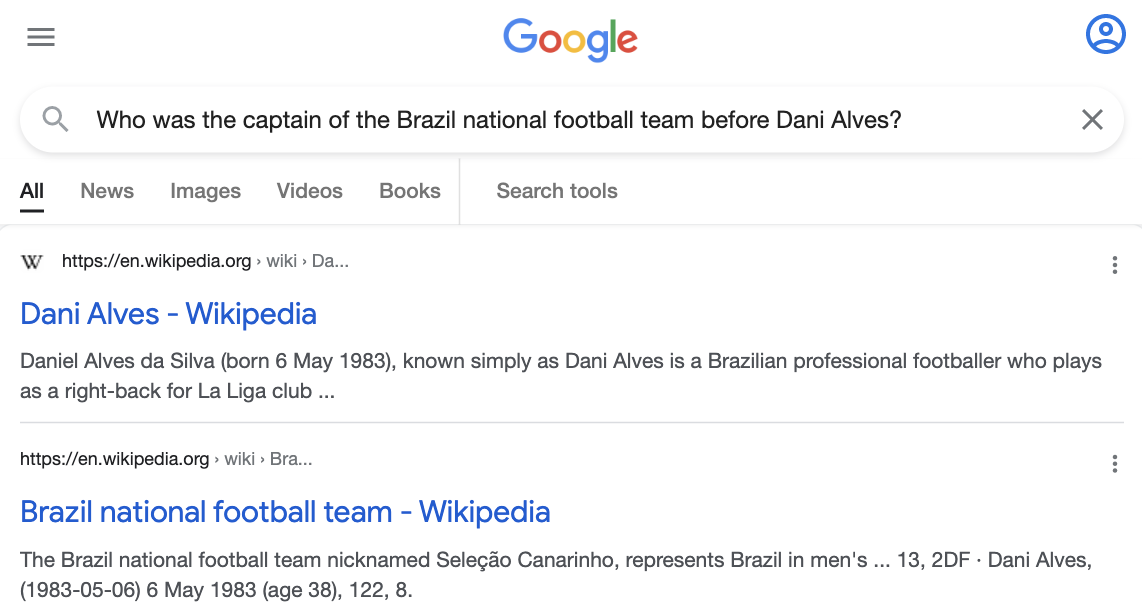}
\caption{\label{fig:example} An example of querying a complex question in Google search. The two search results are wrong, the correct answer is Neymar\protect\footnotemark[2]. The reason for the wrong answer is that it focuses on the entity (\emph{e.g.}, Dani Alves) but fails to capture and utilize the implicit temporal information of ``before Dani Alves".
}
\end{figure}
\footnotetext[2]{Note that the date we conduct the search is March 15, 2022.}

To solve the above problem, the limited literature either decomposes the given question into non-temporal and temporal sub-question to answer \cite{jia2018tequila}, or directly combines the pre-trained language model with the temporal KG to generate answers \cite{saxena2021question}. These methods can achieve satisfying performance on the questions with simple-entity or simple-time (refer to Table~\ref{fig:example} for examples), but fail to answer the questions with complex templates (\emph{e.g.}, using Before/After to construct the question). We argue that the current state-of-the-art methods have not well solved, and may not even be aware of, the following challenges, which we address in this paper:

\textbf{Q1: How to capture the implicit or explicit temporal information in the question to specify the question representation?}
Most of the existing methods directly feed the question into the pre-trained language models to obtain the question embeddings. These approaches would over-rely on information about the entity involved in the given question and ignore the temporal constraint.
Take the question depicted in Fig.~\ref{fig:example} as an example.
The Google search engine might ignore the time constraint ``before Dani Alves'' and purely regard ``Dani Alves'' as the query, which leads to the wrong answers.

\textbf{Q2: How to effectively incorporate the relevant knowledge of temporal KGs into the question representation?}
There is rich temporal information in temporal KGs, which can promote understanding the given question. 
For instance, for the question in Fig.~\ref{fig:example}, one can extract the quadruple (Dani Alves, position held, captain of Brazil, 2019, 2022) in temporal KGs using the entity Dani Alves as the query.
Unfortunately, many prior researches use KGs solely for querying the answer rather than enriching the question representations.

To this end, we propose a \textbf{T}ime-aware \textbf{M}ultiway \textbf{A}daptive (\textbf{TMA}) fusion network for temporal KGQA. 
Specifically, for a given question, we first \emph{select} the relevant knowledge (\textit{i.e.}, Subject-Predicate-Object (\textit{abbr}. SPO)) of the entity in the question from temporal KGs, which is capable of dealing with \textbf{(Q1)}. 
We then adopt multiway attention to perform \emph{matching} between the question and the SPOs.
Next, we design an adaptive \emph{fusion} mechanism to incorporate the SPO into the question representations, which allows the question embeddings to encode the relevant knowledge from temporal KGs, corresponding to \textbf{(Q2)}.
Finally, we generate the final \emph{predictions} by feeding the pre-trained temporal KG embeddings together with the question embeddings into a Multi-Layer Perceptron (MLP) module.

The main contributions of this work can be summarized as follows.
First, We systematically discuss the feasibility of explicitly integrating SPO information into the question for solving temporal KGQA and propose a novel framework called time-aware multiway adaptive (TMA) fusion networks.
Second, We develop a new multiway matching module to capture the temporal information in the question, whose outputs are then fed into a novel adaptive fusion module to incorporate the relevant knowledge from the KGs into the question representations.
Finally, Extensive experiments over temporal datasets demonstrate the superiority of our model compared with other competitive methods. It is worth noticing that, on the CronQuestions dataset, the largest temporal KGQA dataset, our Hits@1 and Hits@10 results of TMA on complex questions are improved by 24\% and 10\% compared to the best-performing baselines.

\section{Related work}
\label{sec:related_works}
In this section, we briefly review some related works, \textit{i.e.}, temporal knowledge graph embedding and temporal QA methods.
\subsection{Temporal Knowledge Graph Embedding}
Knowledge graph embedding (KGE) \cite{bordes2013translating} aims to embed entities and relationships into a low-dimensional continuous vector space, thus facilitating downstream tasks like knowledge graph completion \cite{sun2019pullnet}, relation extraction and classification \cite{liu2021deep,saxena2020improving,xue2023dual} and semantic matching \cite{wang2022dabert,song-etal-2022-improving-semantic,liang2019adaptive,liang2019asynchronous}. However, these methods work on non-temporal KGs but are unsuitable for temporal KGs. Recently, several methods have been proposed to shift the learning capabilities of the model to the temporal KGs. In \cite{jiang2016towards}, the authors present an approach that combines the timestamp embeddings with the score function, which is the first attempt to apply TransE \cite{bordes2013translating} to temporal KGs. Later, 
HyTE \cite{dasgupta2018hyte} leverages time information in the entity-relation space by assigning a corresponding hyperplane to each timestamp. Afterwards, TCompIEx \cite{lacroix2020tensor} then uses the solution based on the canonical decomposition of tensors to further extend the CompIEx \cite{trouillon2016complex}.
\subsection{Temporal QA Methods}
Recently, several approaches \cite{jia2018tequila,jia2021complex} have been proposed to solve this task. TEQUILA \cite{jia2018tequila} decomposes and rewrites each question into temporal and non-temporal sub-questions and then adopts constrained reasoning about time intervals to obtain the desired answers. Moreover, this work also presents a dataset (TempQuestions) dedicated to the temporal KGQA, where the KG used in this dataset is derived from Freebase. Then, EXAQT \cite{jia2021complex}, as the first end-to-end temporal QA system, extracts question-specific subgraphs from the KG and employs relational graph convolutional networks to obtain the updated entity and relation embeddings. Remarkably, to further promote the field of temporal KGQA, a temporal QA question dataset called CronQuestions \cite{saxena2021question} has been released, which is more comprehensive than previous benchmarks. Meanwhile, they introduce a model, namely CronKGQA, which combines the temporal KG embeddings and pre-trained language models and achieves relatively satisfying performance compared to other baselines referred to in this work. However, the aforementioned methods either use hand-crafted rules to tackle the temporal questions or only handle simple question reasoning and gain uncompetitive performance when meeting complex questions with temporal constraints. However, our model does not need to deal with hand-crafted rules while still achieving desirable results in reasoning about complex multi-hop questions.

\section{Problem Definition and Background}
\label{sec:task definition}
In this section, we will introduce the definition of this task and relevant background.

\noindent\textbf{Temporal KGQA.} 
Given a natural language question and a temporal KG $\mathcal{G}=(\mathcal{E}, \mathcal{P}, \mathcal{T}, \Upsilon)$, the task of temporal KGQA is to find a suitable entity $e \in \mathcal{E}$ or timestamp $t \in \mathcal{T}$ that can answer the question accurately. Here, $\mathcal{E}$ denotes a set of entities, and $\mathcal{P}$ denotes a set of predicates. $\mathcal{T}$ represents a set of timestamps. $\Upsilon$ contains many facts that are tuples existing in KGs with the form of $(s, p, o, t)$, where $s, o \in \mathcal{E}$ represent subject and object, respectively, $p\in \mathcal{P}$ is the predicate, and $t \in \mathcal{T}$ is the timestamp. 
For example, for the question $\vartheta$ ``Who was the President of Italy in 2008?'', we can transform it into the form $(?, \vartheta, Italy, 2008)$. 


\noindent\textbf{Temporal KG embedding} aims to learn low-dimensional embeddings $e_s, e_p, e_o, e_t \in \mathbb{R}^{d}$ for each $s, o \in \mathcal{E}, p \in \mathcal{P}, t \in \mathcal{T}$. Generally, we can define a score function $\phi(\cdot)$ based on semantic similarity to learn these embeddings. For a valid fact $\upsilon=(s, p, o, t) \in \Upsilon$, it will be scored much higher than an invalid fact $\upsilon^{\prime}=(s^\prime, p^\prime, o^\prime, t^\prime) \notin \Upsilon$ via the function $\phi(\cdot)$. That is, we need to make $\phi(e_s, e_p, e_o, e_t) > \phi(e_{s^\prime}, e_{p^\prime}, e_{o^\prime}, e_{t^\prime})$.

\noindent\textbf{TComplEx} \cite{lacroix2020tensor} is a semantic matching algorithm specific to the temporal KG, which is the extension of ComplEx \cite{trouillon2016complex}. Concretely, it defines entities, relations, and temporal embeddings in the complex space, and its corresponding score function is as in Eq. \ref{score}.
\begin{equation}
    \label{score}
    \phi(e_s, e_p, \hat{e}_o, e_t) = \mathbf{Re}(\langle e_s, e_p \odot e_t, \hat{e}_o\rangle)
\end{equation}

\noindent where $\mathbf{Re}(\cdot)$ represents the real part and $\langle \cdot \rangle$ denotes the multi-linear product operation. Moreover, $\hat{e}_o$ is the conjugate operation, and $e_s, e_p, e_o, e_t \in \mathbb{C}^d$ are the complex-value embeddings. 
TComplEx has become a prevailing method for inferring missing facts due to this learning paradigm. Therefore, we employ it to generate KG embeddings in this work.

\section{Methods}
\label{sec:method}
\begin{figure}
\centering
\includegraphics[width=0.48\textwidth]{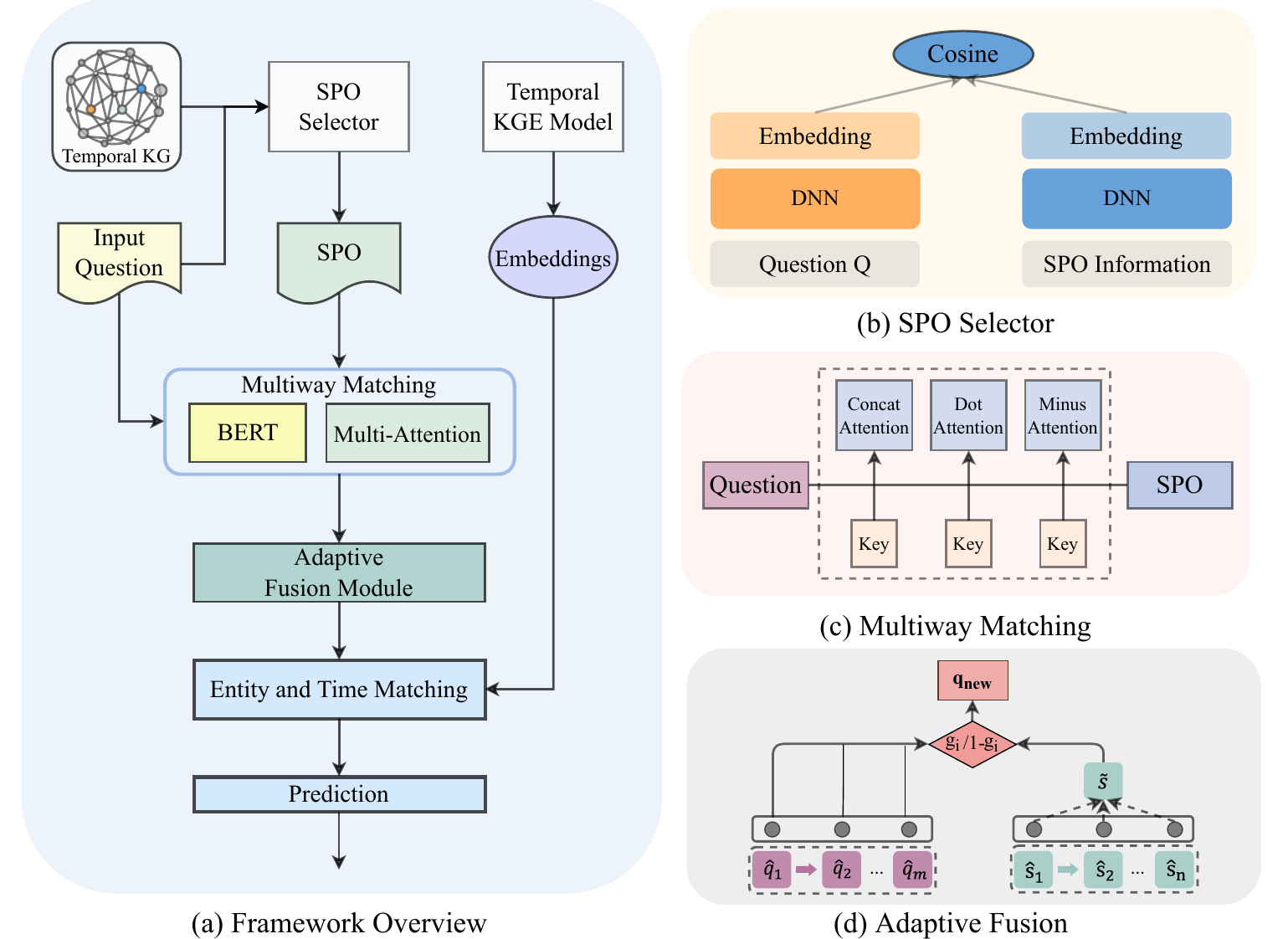}
\caption{\label{fig:model} The overall framework of our model (Best viewed in color).
}
\end{figure}

In this section, we present the details of the proposed TMA. To better understand how it works, its framework is illustrated in Fig. \ref{fig:model}. Concretely, it mainly consists of four essential parts: SPO Selector, Multiway \& Adaptive Fusion, and Answer Prediction. Subsequently, we will elaborate on these critical components that make up our framework.

%

\subsection{SPO Selector}
\label{exp:spo}
We design the SPO selector inspired by Sentence-BERT \cite{reimers2019sentence}, 
As shown in Fig. \ref{fig:model} (b), it is a standard two-tower architecture, where the DNN we use is BERT. 
The tokenized question is fed to the BERT to obtain token embeddings $Q$. SPO information of temporal KG is also performed in the same operations as above, and we can get the SPO embeddings $S$. The concrete formulas are as Eq. \ref{bert}.
\begin{footnotesize}
\begin{equation}
\begin{aligned}
    \label{bert}
    Q &= \textrm{BERT}([\mathrm{CLS}]+question+[\mathrm{SEP}]) \\
    S &= \textrm{BERT}([\mathrm{CLS}]+<\mathrm{SPO}>+[\mathrm{SEP}])
\end{aligned}
\end{equation}
 \end{footnotesize}
\noindent where $Q \in \mathbb{R}^{(n+1) \times d}$ in which $n$ is the number of tokens and $d$ is the dimension of hidden embeddings of the last layer of BERT (\emph{i.e.}, $d=768$). $S \in \mathbb{R}^{c \times d}$ where $c$ is the number of tokens of the SPO. In general, we take the [CLS] embedding (\emph{i.e.}, $q_{[\mathrm{CLS}]}$ and $q_{s_{[\mathrm{CLS}]}}$) as the final question embedding and SPO embedding.

Finally, we apply the cosine similarity on the question and SPO representation to learn the matching scores as follows:
\begin{footnotesize}
\begin{equation}
    \label{cosine}
    score(q_{[\mathrm{CLS}]}, q_{s_{[\mathrm{CLS}]}}) = \frac{q_{[\mathrm{CLS}]}^\top q_{s_{[\mathrm{CLS}]}}}{\Vert q_{[\mathrm{CLS}]}\Vert \Vert q_{s_{[\mathrm{CLS}]} \Vert}}
\end{equation}
\end{footnotesize}
\noindent where $score$ is a scalar. The top ten scored SPOs are selected as candidate information.

\subsection{Multiway \& Adaptive Fusion}
\label{exp:MA}
Previous studies \cite{rocktaschel2015reasoning,tan2018multiway} demonstrate the effectiveness of word-level attention in sentence pair modeling. 
In the multiway attention module, as shown in Fig. \ref{fig:model} (c), 
different attention used to compare the question and SPO  relationship from different perspectives.

For a given question, we embed it with Eq. \ref{bert}, excluding the [CLS] token, \emph{i.e.}, $\bar{Q}=[q_1, q_2, \dots, q_n]$. For the ten selected SPOs, we take the [CLS] token of each SPO and concatenate them together, \emph{i.e.}, $P=[S_1, S_2, \dots, S_{m}]$ ($m$ is the number of selected SPOs). Then, the candidate SPOs can be matched by the word at each position $k$ of the question. 
which are formulated as follows:
\begin{footnotesize}
\begin{equation}
    \tilde{p}_k^\ell = \Phi_\ell(P, q_k;\mathbf{W}_\ell)
\end{equation}
\end{footnotesize}
\noindent where $\tilde{p}_k^\ell$ is the corresponding weighted-sum representation of SPOs specified by $q_k$, employing the attention function $\Phi_\ell$ with parameterized by $\mathbf{W}_\ell$, in which $\ell$ denotes concat attention, dot attention, and minus attention, respectively. More precisely, the different attention mechanisms can be described as follows:

\noindent \textbf{Concat Attention}:
\begin{footnotesize}
\begin{equation}
\begin{aligned}
    h^k_j&=v_{cat}^\top\tanh(\mathbf{W}_{cat}[q_k, S_j]) \\
    \alpha_i^k&=\exp(h^k_i)/\sum\nolimits_{j=1}^m\exp(h^k_j) ,\quad \tilde{p}_k^{cat}=\sum\nolimits_{i=1}^m\alpha_i^kS_i
\end{aligned}
\end{equation}
\end{footnotesize}

\noindent \textbf{Dot Attention}:
\begin{footnotesize}
\begin{equation}
\begin{aligned}
    h^k_j&=v_{dot}^\top\tanh(\mathbf{W}_{dot}(q_k \odot S_j)) \\ \alpha_i^k&=\exp(h^k_i)/\sum\nolimits_{j=1}^m\exp(h^k_j),\quad \tilde{p}_k^{dot}=\sum\nolimits_{i=1}^m\alpha_i^kS_i
\end{aligned}
\end{equation}
\end{footnotesize}

\noindent \textbf{Minus Attention}:
\begin{footnotesize}
\begin{equation}
\begin{aligned}
    h^k_j&=v_{min}^\top\tanh(\mathbf{W}_{min}(q_k - S_j))\\ \alpha_i^k&=\exp(h^k_i)/\sum\nolimits_{j=1}^m\exp(h^k_j),\quad \tilde{p}_k^{min}=\sum\nolimits_{i=1}^m\alpha_i^kS_i
\end{aligned}
\end{equation}
\end{footnotesize}


Next, to obtain the attention-based question representation $\tilde{Q}^\ell$, we aggregate the matching information $\tilde{p}_k^\ell$ together with the word representation $q_k$ via the concatenation operation, \textit{i.e.}, $\tilde{q}_k^\ell=[q_k, \tilde{p}_k^\ell]$. Finally, the linear transformation is applied to $\tilde{Q}^\ell$ that fuses the SPO information, \emph{i.e.}, $\mathcal{Q}_{final}=\mathbf{W}[\tilde{Q}^{cat}, \tilde{Q}^{dot}, \tilde{Q}^{min}]=[\hat{q}_1, \dots, \hat{q}_n]$. Similarly, the question can be matched by a particular SPO by performing the above multiway operation and linear transformation. In this way, we can obtain the updated SPO representation $\hat{S}_i$.

\noindent \textbf{Adaptive Fusion}: To make question representations more time-aware, as shown in Fig. \ref{fig:model} (d), we use a gate mechanism to adaptively fuse the temporal information from SPOs.
\begin{footnotesize}
\begin{equation}
    \begin{aligned}
        \tilde{S} &= \tanh(\mathbf{W}_{\hat{S}_i}\frac{1}{m}\sum\nolimits_{i=1}^m\hat{S}_i+b_{\hat{S}_i}) \\
        g_i &= \sigma(\mathbf{W}_{g_i}(\hat{q}_i\cdot\tilde{S})) ,\quad q_{\mathrm{new}}=g_i\hat{q}_i + (1-g_i)\tilde{S}
    \end{aligned}
\end{equation}
\end{footnotesize}
\noindent where $\sigma$ denote the nonlinear activation function, respectively. $q_{\mathrm{new}}$ is the final embedding of the word in the question.
\subsection{Answer Prediction}
\label{exp:AP}
First, we obtain two embeddings by projecting $q_{\text{new}}$, $q_{ent}$ and $q_{time}$, which are specific to entity and time, respectively. Then, we get the entity score and timestamp score by the entity scoring function and the time scoring function, respectively.

\noindent \textbf{Entity scoring function}: As mentioned in Section~\ref{sec:task definition}, we calculate each entity $\hat{e} \in \mathcal{E}$ score by using the score function $\phi(\cdot)$ as
\begin{footnotesize}
\begin{equation}
    \phi_{ent}(\hat{e}) = \textbf{Re}(\langle e_s, q_{ent} \odot e_t, \hat{e}\rangle)
\end{equation}
\end{footnotesize}
\noindent where $s$ and $t$ are the entity and the timestamp extracted from the question. $e_s$ and $e_t$ are the corresponding embeddings computed by the pre-trained TComplEx method. Note that if $s$ and $t$ do not exist, we use a dummy entity instead.

\noindent \textbf{Time scoring function}: We first extract the subject $s$ and the object $o$ from a given question. Then each timestamp $\hat{t} \in \mathcal{T}$ score can be computed as follows.
\begin{footnotesize}
\begin{equation}
        \phi_{time}(\hat{t}) = \textbf{Re}(\langle e_s, q_{time} \odot e_{\hat{t}}, e_o\rangle)
\end{equation}
\end{footnotesize}
Finally, the obtained score for all entities and timestamps are concatenated, and the answer probabilities are computed with a softmax layer over these combined scores. The objective function we adopted is cross-entropy loss, which is defined as.
\begin{footnotesize}
\begin{equation}
    \mathcal{L}=-\sum_i^Ny_i\log(\hat{y}_i)+(1-y_i)\log(1-\hat{y}_i)
\end{equation}
\end{footnotesize}
\noindent where $y_i$ is the ground truth label and $\hat{y}_i$ is the predicted label.

\begin{table*}[h]
	\centering
 \renewcommand\arraystretch{1.0}
	\caption{Performance of baselines and our methods on the CronQuestions dataset.}
	\label{results} 
	\scalebox{0.85}{
        \setlength{\tabcolsep}{4.3mm}{
	\begin{tabular}{l|c|cc|cc|c|cc|cc}
	    \hline
		\multirow{3}*{\textbf{Model}}& \multicolumn{5}{c|}{\textbf{Hits@1}} & \multicolumn{5}{c}{\textbf{Hits@10}} \\
		\cline{2-11}
        ~& \multirow{2}*{\textbf{Overall}} & \multicolumn{2}{c|}{\textbf{Question Type}} & \multicolumn{2}{c|}{\textbf{Answer Type}} & \multirow{2}*{\textbf{Overall}}& \multicolumn{2}{c|}{\textbf{Question Type}} & \multicolumn{2}{c}{\textbf{Answer Type}} \\
        \cline{3-6}\cline{8-11}
        ~&~& \textbf{Complex} & \textbf{Simple} & \textbf{Entity}&\textbf{Time}&~&\textbf{Complex} & \textbf{Simple} & \textbf{Entity}&\textbf{Time}\\
        \hline
        BERT & 0.071 & 0.086 & 0.052 & 0.077 & 0.06 & 0.213 & 0.205 & 0.225 & 0.192 & 0.253 \\
        RoBERTa & 0.07 & 0.086 & 0.05 & 0.082 & 0.048 & 0.202 & 0.192 & 0.215 & 0.186 & 0.231 \\
        KnowBERT & 0.07 & 0.083 & 0.051 & 0.081 & 0.048 & 0.201 & 0.189 & 0.217 & 0.185 &0.23\\
        T5-3B&0.081 & 0.073 &0.091 & 0.088 & 0.067&-&-&-&-&- \\
        \hline
        EmbedKGQA & 0.288 & 0.286 & 0.29 & 0.411 & 0.057 & 0.672 & 0.632 & 0.725 & 0.85& 0.341\\
        T-EaE-add& 0.278 & 0.257 & 0.306 & 0.313& 0.213& 0.663 &0.614 & 0.729 & 0.662 & 0.665\\
        T-EaE-replace & 0.288 & 0.257 & 0.329 & 0.318 & 0.231 & 0.678 & 0.623 & 0.753 & 0.668 & 0.698\\
        CronKGQA & 0.647 & 0.392 & 0.987 &0.699 & 0.549 & 0.884&0.802&0.992 & 0.898 & 0.857\\
        \hline
        TMA (ours) & \textbf{0.784} & \textbf{0.632} & \textbf{0.987} & \textbf{0.792} & \textbf{0.743} &\textbf{0.943} &\textbf{0.904} &\textbf{0.995} &\textbf{0.947} &\textbf{0.936}\\ \hline
	\end{tabular}}}
\end{table*}

\section{Experiments}
\label{sec:exp}
\textbf{Dataset.} To validate the effectiveness of our proposed model for the temporal KGQA task, we employ a benchmark dataset, called CronQuestions, that has been widely used in the previous study \cite{saxena2021question}. 


\noindent \textbf{Baselines.} To demonstrate the superiority of TMA, we select two types of baseline models for comparison. (I) Pre-trained language models includes BERT, RoBERTa, T5 and KnowBERT. (II) KG embedding-based approaches includes the variants of Entities as Experts (\textit{i.e.}, T-EaE-add and T-EaE-replace), EmbedKGQA \cite{saxena2020improving} and CronKGQA \cite{saxena2021question}. 

\section{Results and analysis}
\subsection{Model Performance}
\label{res:MP}

We evaluate the performance of TMA and other competitive models on the temporal KGQA dataset, \emph{i.e.}, CronQuestions. Table \ref{results} shows the results of different baselines in terms of Hits@1 and Hits@10 across different question types and answer types. Our proposed model achieves state-of-the-art performance on all types of questions and answers on Hits@1 and Hits@10, which illustrates its powerful representation capability. Remarkably, the Hits@1 and Hits@10 results of TMA on complex questions are improved by 24\% and 10\%, respectively, compared to the second best-performing model. One plausible reason is that complex reasoning requires a better understanding of the question representation, and TMA can incorporate relevant temporal-specific information from the KG due to the design of SPO selector and multiway \& adaptive fusion.

Furthermore, we find that the performance of methods based only on large-scale pre-trained language models (\emph{i.e.}, BERT, RoBERTa, T5, and KnowBERT) is significantly worse than that of KG embedding-based methods (\emph{i.e.}, EmbedkGQA, T-EaE, and CronKGQA). This indicates that it is difficult to capture the time-aware question embeddings through pre-trained language models alone. 
Meanwhile, T5 is better than other pre-trained language models since it contains more trainable parameters. 

\begin{table}[h]
	\centering
	\caption{Hits@1 for different reasoning type questions.}
	\label{category}
	\resizebox{0.45\textwidth}{!}{
	\begin{tabular}{l|c c c|c c|c}
	    \hline
	    \multirow{3}*{\textbf{Category}}&\multicolumn{3}{c|}{\textbf{Complex Question}}&\multicolumn{2}{c|}{\textbf{Simple Question}}&\multirow{3}*{\textbf{All}}\\
	    \cline{2-6}
        ~&\textbf{Before/}&\textbf{First/}&\textbf{Time}&\textbf{Simple}&\textbf{Simple}&~ \\
        ~&\textbf{After} & \textbf{Last} & \textbf{Join} & \textbf{Entity}& \textbf{Time} & ~ \\
        \hline
        EmbedKGQA & 0.199 & 0.324 & 0.223 & 0.421 & 0.087 & 0.288 \\
        T-EaE-add& 0.256 & 0.285 & 0.175 & 0.296 & 0.321 & 0.278\\ 
        T-EaE-replace & 0.256 & 0.288 & 0.168 & 0.318 & 0.346 & 0.288\\ 
        CronKGQA&0.288 & 0.371 &0.511 & 0.988 & 0.985&0.647 \\
        TMA (ours)&\textbf{0.581} &\textbf{0.627}&\textbf{0.675}&\textbf{0.988}&\textbf{0.987}&\textbf{0.784} \\ 
        \hline
	\end{tabular}
	}
\end{table}

\subsection{Performance across question categories}
\label{res:Pa}

We compare our model against various KG-based methods with different categories of questions in terms of Hits@1, and summarize the results in Table \ref{category}. As the table depicts, the proposed method TMA obtains better performance than all the KG-based methods, especially on those complex questions. TMA's performance in Before/After", First/Last", and Time Join" can consistently achieve 30\%, 25\%, and 16\% improvements, respectively, when compared to the CronKGQA. 
This phenomenon 
further verifies our idea of explicitly incorporating the SPO information into the question to learn temporal-specific question representations for temporal KGQA tasks. 

\begin{table}[h]
	\centering
	\caption{Results of component ablation experiment.}
	\label{ablation}  
	\scalebox{0.8}{
	\begin{tabular}{l|c|c c|c c}
	    \hline
	    \multirow{3}*{\textbf{Model}}& \multicolumn{5}{c}{\textbf{Hits@1}}\\
		\cline{2-6}
        ~& \multirow{2}*{\textbf{Overall}} & \multicolumn{2}{c|}{\textbf{Question Type}} & \multicolumn{2}{c}{\textbf{Answer Type}} \\
        \cline{3-6}
        ~&~& \textbf{Complex} & \textbf{Simple} & \textbf{Entity}&\textbf{Time}\\
        \hline
        TMA &0.784&0.632&0.987&0.792&0.743\\
        w/o SPO Selector&0.726&0.584&0.916&0.736&0.707\\
        w/o Concat Attention&0.759&0.628&0.934&0.769&0.739\\
        w/o Dot Attention&0.745&0.617&0.914&0.771&0.732\\
        w/o Minus Attention&0.768&0.630&0.952&0.789&0.728\\
        w/o Adaptive Fusion&0.736&0.613&0.899&0.742&0.724\\
        \hline
	\end{tabular}
	}
\end{table}

\subsection{Ablation Study}
\label{res:AS}
To evaluate the contribution of each module in our framework, we perform extensive ablation experiments. The experimental results are shown in Table \ref{ablation}. 

The SPO Selector can select the SPO triples from the temporal KG relevant to the semantics of the question. When we remove the SPO Selector, the performance drops to 0.726, indicating that the candidate SPOs are critical for this task.

The multiway attention module is composed of three components. When we remove concatenation attention, dot attention, and minus attention, the performance drops to 0.759, 0.745, and 0.768, respectively. The concatenation attention, which is often employed in retrieving QA, is significantly improved for ``Entity'' questions. This is probably because the concatenation attention facilitates the fusion between entities and provides more detailed entity alignment information. In addition, minus attention significantly improves on the ``Time'' question. The possible reason is that it can explicitly align the differences between entities and time, thus providing better underlying features for adaptive fusion. 

Finally, we remove the adaptive module for information fusion, which is equivalent to directly using the SPO information and the original semantic information fusion. The performance of TMA on simple reasoning drops by nearly 9\%, which indicates that adaptive fusion can effectively integrate two different kinds of information.

\section{Conclusion}
Temporal KGQA task exists a problem that is not capable of learning \emph{temporal-specific} embeddings of entities. We propose a method called \textbf{TMA}, 
which can explicitly fuse the SPO to question representations by a \emph{select-match-fusion-predict} paradigm. This method improves the model's robustness by enabling the obtained question embeddings to be more temporal-specific. Extensive experiments on CronQuestions dataset verify the effectiveness of the TMA. 
\label{sec:conclusion}


\bibliographystyle{IEEEtran}
\bibliography{strings,refs}

\end{document}